# Knowledge-aware Attention Network for Protein-Protein Interaction Extraction

Huiwei Zhou[1,*], Zhuang Liu[1], Shixian Ning[1], Chengkun Lang[1], Yingyu Lin[2] and Lei Du[3]

[1]School of Computer Science and Technology, Dalian University of Technology, Dalian 116024, Liaoning, China

[2]School of Foreign Languages, Dalian University of Technology, Dalian, China 116024, Liaoning, China

[3]School of Mathematical Sciences, Dalian University of Technology, Dalian, China 116024, Liaoning, China

[*] Corresponding author.

Email: zhouhuiwei@dlut.edu.cn, {zhuangliu1992, ningshixian, kunkun}@mail.dlut.edu.cn, {lyydut, dulei}@dlut.edu.cn

**Abstract** Protein-protein interaction (PPI) extraction from published scientific literature provides additional support for precision medicine efforts. However, many of the current PPI extraction methods need extensive feature engineering and cannot make full use of the prior knowledge in knowledge bases (KB). KBs contain huge amounts of structured information about entities and relationships, therefore plays a pivotal role in PPI extraction. This paper proposes a knowledge-aware attention network (KAN) to fuse prior knowledge about protein-protein pairs and context information for PPI extraction. The proposed model first adopts a diagonal-disabled multi-head attention mechanism to encode context sequence along with knowledge representations learned from KB. Then a novel multi-dimensional attention mechanism is used to select the features that can best describe the encoded context. Experiment results on the BioCreative VI PPI dataset show that the proposed approach could acquire knowledge-aware dependencies between different words in a sequence and lead to a new state-of-the-art performance.

## 1 Introduction

The intricate networks of protein-protein interactions (PPIs) contribute to controlling cellular homeostasis and the development of diseases in specific contexts. Understanding how gene mutations and variations affect the cellular interactions provides vital support for precision medicine efforts. Although numerous PPIs are manually curated into structure knowledge databases (KB) by biomedical curators, such as IntAct [1] and BioGrid [2], many valuable PPIs remain available in the growing amount of scientific articles. However, manually extracting these PPIs from biomedical literature is expensive and difficult to keep up-to-date. Automatically extracting these relations from even increasing volumes of scientific literature is of great importance to expediting database curation.

To promote these issues, the BioCreative VI has proposed a challenging task of applying text-mining methods to automatically extracting interaction relations of protein-protein pairs affected by genetic mutations. The novel challenge for the biomedical natural language processing community generates a lot of interest and participation. Many automatic PPI extraction methods have been proposed, which can be divided into three categories: rule-based methods, feature-based methods and neural network-based methods.

Rule-based [3] methods are simple and effective, but hard to apply to a new dataset. Feature-based methods extract PPIs or drug-drug interactions based on one-hot represented lexical and syntactic features [4], [5], [6], [7]. Generally, the PPIs extraction performance relies heavily on the suitable features, which require extensive feature engineering.

Recently, neural network-based methods have been proposed to map word and entity sequences into a low-dimensional vector space, and then learn semantic representations of word sequences for relation extraction without making many feature engineering efforts. Zeng et al. [8] first employ Convolutional Neural Network (CNN) [9] to learn sentence-level representations for relation extraction and it achieves better performance than feature-based methods. As for BioCreative VI PPI extraction task, Tran and Kavuluru [10] employ CNN to extract local semantic features and get 30.11% F1-score. CNN pays more attention to local features by performing convolutions within the varying filter windows. This hierarchical structure is good at presenting local or position-invariant features, but neglects the long-range dependencies.

Some efforts have been put into capturing long-term structure within sequences by using recurrent neural network (RNN) [11], long short-term memory network (LSTM) [12] and memory network models [13], [14]. For PPI extraction task, Wang et al. [15] propose a RNN-based method, which makes full use of word, entity and sentence information through attention mechanisms. Zhou et al. [16] use LSTM to model long-distance relation patterns for chemical disease relation extraction. Sahu and Anand [17] propose a joint bidirectional LSTM model with word and position embedding to extract drug-drug interaction relations.

Zhou et al. [18] adopt memory networks for PPI extraction, and show that an external memory are superior to long short-term memory networks with the local memories. However, the memory networks lack the capacity to capture long-range dependencies.

Recently, attention mechanism is used extensively to enhance the representation capability of CNN or RNN, and has succeeded in various NLP tasks [19], [20], [21]. The attention mechanism uses a hidden layer to compute the weight/importance of each element in the input sequence and then defines sequence representations as the weighted sum of element representations. It can thus capture long-range dependencies. However, such attention mechanisms are used in conjunction with CNN or RNN.

To eschew recurrence and convolution, a sole attention mechanism, called "Transformer", is proposed to construct a sequence to a sequence model which achieves a state-of-the-art performance in the neural machine translation task [20]. Transformer relies entirely on self-attention and multi-head attention to compute dependencies between any two context words, which makes it easier to learn long-range dependencies. However, it has not been used for relation classification task.

On the other hand, large-scale KBs usually store prior knowledge in the form of triplet (head entity, relation, tail entity), (also denoted as (h, r, t)). The relation indicates the relationship between the two entities. This prior knowledge driven from KBs is very effective in relation extraction. Some researchers [22], [23] derive prior knowledge features from knowledge bases, such as Wikipedia and biological knowledge bases, to enhance their models. However, these methods describe knowledge features as one-hot representations, which assume that all the objects are independent from each other. Recently, knowledge representation learning methods have been proposed for encoding the entities and relations into low-dimensional vector space and could find the potential semantic relations between entities and relations. Among these methods, TransE [24] is simple but can achieve the state-of-the-art predictive performance.

To explicitly capture long-range dependencies and introduce the prior knowledge for PPIs extraction, this paper proposes a novel knowledge-aware attention network (KAN) for PPI extraction without using RNN and CNN. Our proposed KAN employs TransE to learn embeddings of protein entities and relations from KBs, which are then used to capture important context representations with two levels of attention: diagonal-disable multi-head attention (DMHA) and multi-dimensional attention (MDA). For each word of the input sequence, DMHA calculates its semantic relatedness with all words in the sequence with the guide of entity embeddings. In this way, a sequence of vectors which contain knowledge-aware dependencies between different words are acquired. And then for each vector of the sequence, MDA computes a weighted score for each feature in the vector to select the important features. Compared with RNN\CNN, our attention mechanism could learn long-range dependencies of a sequence and conduct direct connections between two arbitrary tokens in a sequence. Meanwhile, the prior knowledge is integrated into the context representations. Experiment on the BioCreative VI PPI dataset [25] shows that the proposed approach leads to a new state-of-the-art performance.

## 2 Background

2.1 Knowledge Representation Learning

In order to embed both the entities and relations into a continuous vector space, a variety of knowledge representation learning methods [24], [26], [27], [28], [29] have been developed. TransE is a typical knowledge representation approach, which represents the relation between the two entities as a translation in a representation space, that is, $\mathbf{h+r \approx t}$ when (h, r, t) holds. TransE is simple and efficient, and it can achieve state-of-the-art performance on modeling KBs [24]. This paper employs TransE to learn embeddings of protein entities and relations from KBs.

2.2 Self-attention

Self-attention could relate different positions of a single sequence by computing the attention score with each pair of tokens in the sequence. More formally, self-attention aims at mapping a query and a set of key-value pairs to an output, where the query ($Q$), keys ($K$), values ($V$) and output are all vectors. In self-attention, the $Q$, $K$ and $V$ are the same sequence. Self-attention uses a dot product to calculate the semantic relatedness of $Q$ with $K$ and applies a softmax function to obtain the attention weight on the $V$. The output of self-attention is computed as:

$$\text{Attention}(Q,K,V) = \text{softmax}(\frac{QK^T}{\sqrt{d_k}})V \quad (1)$$

where $d_k$ is the dimension of $Q$ and $K$. Self-attention could capture the long-range dependencies without any RNN/CNN structure. Self-attention has been used successfully in a many tasks including reading comprehension, abstractive summarization, textual entailment and learning task-independent sentence representations [30], [31], [32], [33]. This paper introduces a novel DMHA based on self-attention for better capturing the context representations and the knowledge representations.

## 3 Method

The PPI extraction task proposed by BioCreative VI [34] aims at automatically extracting protein-protein interaction relations affected by genetic mutations (PPIm). Our method uses the prior knowledge and contextual information of two protein entities to identity whether they participate in a PPIm relationship. This section first describes the preprocessing procedure and then introduces the feature representation to ease the exposition. Finally, KAN is described in detail.

3.1 Preprocessing and Feature Representation

Since PPIm relations on the BioCreative VI PPI dataset [25] are annotated at the document level, the context word sequences of each protein pair in a document are considered as candidate instances. To reduce the number of inappropriate instances, the sentence distance between a protein pair should be less than 3. Otherwise, the protein pair will not be considered. We select the words between a protein pair and three expansion words on both sides as the context word sequence with respect to the protein pair. To simplify the interpretation, we consider the mentions of a protein pair as two single words $C_{i_1}$ and $C_{i_2}$, where $i_1$ and $i_2$ are the positions of the protein pair. For a given text $\{...,c_1,c_2,c_3,c_{i_1},c_{i_1+1}...,c_i,...,c_{i_2-1},c_{i_2},c_{L-2},c_{L-1},c_L,...\}$, the context word sequence generated can be expressed as $\{c_1,c_2,c_3,c_{i_1+1}...,c_i,...,c_{i_2-1},c_{L-2},c_{L-1},c_L\}$. As can be seen, we remove the mentions of the protein pair to be classified in the current instance. Then all the other protein mentions are replaced with "gene0". The numbers in the context are replaced by a specific string "NUMBER". Some special characters, such as "*", are removed. In the training phase, if a protein pair is annotated as an interacting pair in a document, each context sequence of the protein pair in the document is recognized as a positive instance, otherwise a negative instance. In the test phase, a protein pair is recognized as an interacting pair as long as one of its instances is classified as positive.

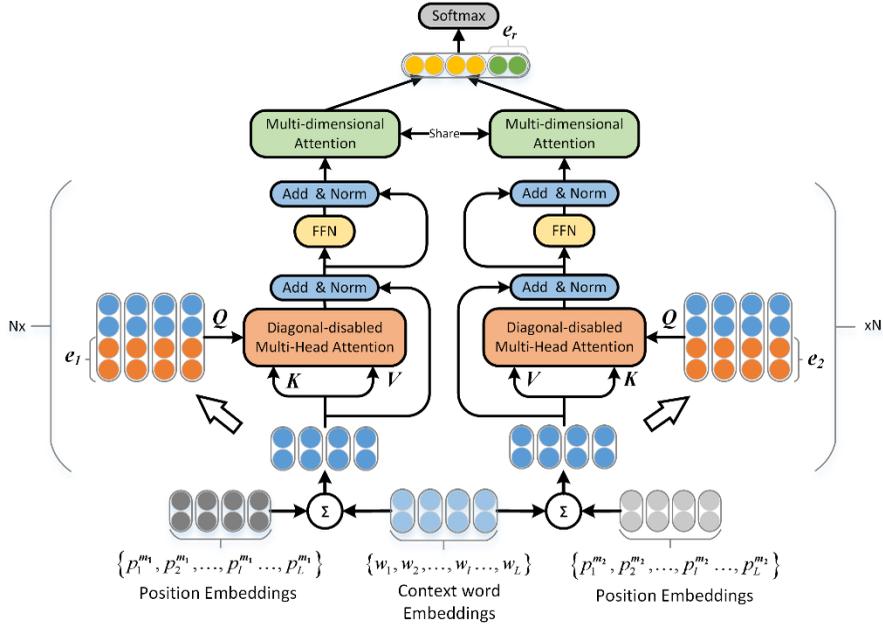

Fig.1. The structure of the knowledge-aware attention network. It has two components with the same structure. Each component contains a diagonal-disabled multi-head attention (DMHA) layer, a feed-forward networks (FFN) layer and a multi-dimensional attention (MDA) layer. The DMHA and FFN layers would be repeated multiple times to improve the performance. The two components share the same MDA layer.

The input to our model consists of two parts, the context word sequence and prior knowledge of the protein pair. We transform the words in the context sequence into continuous vectors, word embeddings. Word embeddings preserve both syntactic and semantic information well [35]. Thus, the context of the protein pair is represented as a sequence of word embeddings $\{w_1, w_2, ..., w_l, ..., w_L\}$, where $w_l \in \mathbb{R}^d$ and $L$ is the length of the context.

The entity-relation triples in KBs are considered as prior knowledge. TransE is selected to embed protein-protein relation triples $(m_1, r, m_2)$ into a continuous vector space $\mathbb{R}^d$, represented as $(e_1, e_r, e_2)$ where the relation embeddings $e_r$ corresponds to a translation between the embeddings of the protein pair $(e_1, e_2)$.

Furthermore, we incorporate word position embeddings to reflect the relative distance from the words to two protein mentions. Word position embeddings have been shown to improve the performance of relation classification [8]. The position of each context word to one protein is defined as the absolute distance from the context word to the protein mention. Since there are two entities $(m_1, m_2)$, two sequences of position embeddings are obtained, which are represented as $\{p_1^{m_1}, p_2^{m_1}, ..., p_l^{m_1}, ..., p_L^{m_1}\}$ and $\{p_1^{m_2}, p_2^{m_2}, ..., p_l^{m_2}, ..., p_L^{m_2}\}$, respectively. The position embeddings have a dimension size of $d$, same as word embeddings. Following Vaswani et al. [20], we encode the position and add the position embedding to the context word embedding which can be formalized as follows:

$$x_l^{m_i} = w_l + p_l^{m_i} \qquad (2)$$

where $i \in \{1, 2\}$ and $m_i$ represents one of the two protein entities. Thus, we obtain two sequences of context word embeddings with position information, $\{x_1^{m_1}, x_2^{m_1}, ..., x_l^{m_1}, ..., x_L^{m_1}\}$ and $\{x_1^{m_2}, x_2^{m_2}, ..., x_l^{m_2}, ..., x_L^{m_2}\}$. We list important notations in the supplementary material.

3.2 Knowledge-aware Attention Network (KAN)

This section describes the details of KAN. The overall structure of KAN is shown in Fig. 1, which has two components with respect to two protein entities. Two sequences $\{x_1^{m_1}, x_2^{m_1}, ..., x_l^{m_1}, ..., x_L^{m_1}\}$ and $\{x_1^{m_2}, x_2^{m_2}, ..., x_l^{m_2}, ..., x_L^{m_2}\}$ along with two entity embeddings $e_1$ and $e_2$ are the inputs of the two components, respectively. Two components have the same structure which consists of a diagonal-disabled multi-head attention (DMHA) layer, a feed-forward networks (FFN) layer and a multi-dimensional attention (MDA) layer. Firstly, the DMHA layer and FFN layer would be repeated multiple times (2 times in this paper for the low complexity) to improve the performance. We anticipate that DMHA and FFN could conduct direct connections between two arbitrary tokens and learn long-range dependencies of the context. Then, a shared multi-dimensional attention (MDA) layer is adopted to compute feature-wise scores to select important features that can best describe the context sequences. Finally, knowledge-aware context representations are obtained, which are concatenated with relation embedding for relation extraction.

Since two components have the same structure, we take one component with respect to one of the protein pair, say $m_1$, for example in the following subsections.

*3.2.1 Diagonal-disabled Multi-head Attention (DMHA)*

Self-attention is an attention mechanism relating different positions of a single sequence by computing the attention weights between each pair of tokens. It is very expressive and flexible for long-range dependencies [21]. After all, PPI is often reflected by complex global semantic information. The diagonal-disabled self-attention mechanism of DMHA is as follows:

$$g_{i,j} = \frac{Q_i K_j^T}{\sqrt{d_a}} + M_{i,j}$$

$$\alpha_{i,j} = \frac{\exp(g_{i,j})}{\sum_{l=1}^{L}\exp(g_{i,l})}$$

$$\text{Att}_i = \sum_{j=1}^{L} \alpha_{i,j} V_j \quad (3)$$

where $L$ is the length of sequence. Self-attention first computes the dot production $g_{i,j}$ between each row of $Q$ (denoted as $Q_i$) and each row of $K$ (denoted as $K_j$). Then, the results are passed to a softmax operation to get attention weights $\alpha_{i,j}$. Finally, the weighted sum of the row vectors of $V$ (denoted as $V_j$) is considered as a row vector $\text{Att}_i$ and stacked into a matrix $\text{Att}(Q,K,V) = [\text{Att}_1, \text{Att}_2, ... \text{Att}_L]$. $M$ is a diagonal-disabled mask matrix, where the diagonal elements $M_{i,i}$ are $-\infty$ and the other elements $M_{i,j}$ are zero. The matrix $M$ aims to disable the attention of each token to itself, which has successfully applied in the Natural Language Inference task [21]. $1/\sqrt{d_a}$ is a scaling factor. Vaswani et al. [20] point out that for large values of $d_a$, the dot products grow large in magnitude, pushing the softmax function into regions where it has extremely small gradients. Therefore, the dot productions are scaled by $1/\sqrt{d_a}$ to counteract this effect.

Vaswani et al. [20] indicates that instead of performing a single attention function with $d$-dimension $Q$, $K$ and $V$, it is beneficial to linearly project $Q$, $K$ and $V$ $h$ times, with different linear projections to $d_q$, $d_k$ and $d_v$ dimensions, respectively. Thus, DMHA performs the diagonal-disabled self-attention multiple times on linearly projected $Q$, $K$ and $V$, which is called multi-head. The multi-head self-attention allows the model to jointly attend to information from different representation subspaces at different positions [20]. DMHA is formalized as follows:

$$\text{DMHA}(Q,V,K) = [head_1, head_2, \cdots, head_h]W^H$$
$$head_i = \text{Att}(QW_i^Q, KW_i^K, VW_i^V) \quad (4)$$

where $W^H \in \mathbb{R}^{hd_{head} \times d}$, $W^Q \in \mathbb{R}^{2d \times d_{head}}$, $W^K \in \mathbb{R}^{d \times d_{head}}$, $W^V \in \mathbb{R}^{d \times d_{head}}$ are the learned parameters matrices, $h$ is the number of heads, and $d_{head} = d/h$ is the dimension of the projected $Q$, $K$ and $V$. $K$ and $V$ in DMHA are both $\{x_1^{m_1}, x_2^{m_1}, ..., x_l^{m_1}, ..., x_L^{m_1}\}$. As for $Q$, we further concatenate the corresponding protein embedding $e_1$ to each position, learned from KBs. Hence, $Q$ can be represented as $\{[x_1^{m_1}, e_1], [x_2^{m_1}, e_1], ..., [x_l^{m_1}, e_1]..., [x_L^{m_1}, e_1]\}$. In this way, we expect to effectively encode the prior knowledge of the protein embedding for each sequence word and aggregate evidence relevant to the current protein from every word in the sequence.

In addition, we employ a residual connection [36] on DMHA layer, followed by a normalization [37] layer:

$$V^{DMHA} = \text{LayerNorm}(V + \text{DMHA}(Q,V,K)) \quad (5)$$

*3.2.2 Feed-Forward Networks (FFN)*

The output vectors $V^{DMHA}$ of the DMHA layer are fed into fully connected Feed-Forward Networks (FFN) layer which consists of two linear transformations with a rectified linear unit (ReLU) activation function.

$$\text{FFN}(V^{DMHA}) = \text{ReLU}(V^{DMHA}W_1 + b_1)W_2 + b_2 \quad (6)$$

where $W_1$, $b_1$, $W_2$ and $b_2$ are the learned parameters. In the same way as Transformer, we employ a residual connection [36] on FFN layer, followed by normalization [37] layer:

$$V^{FFN} = \text{LayerNorm}(V^{DMHA} + \text{FFN}(V^{DMHA})) \quad (7)$$

As depicted at the beginning of this section, DMHA and FFN layers are repeated multiple times, which means that $V^{FFN}$ is V and K of the next DMHA layer and the concatenation of $V^{FFN}$ and entity embeddings forms $Q$. To ease the exposition, we still use $V^{FFN}$ to represent the output vector of the last FFN layer in the following subsections.

*3.2.3 Multi-dimensional Attention (MDA)*

To extract the important features from $V^{FFN}$, a multi-dimensional attention is employed to compresses the sequence of vectors into a vector representation $s$.

$$A = \text{softmax}(\tanh(V^{FFN}W^{att} + b^{att}))$$

$$s = \sum_{i=1}^{L} A_i \odot V_i^{FFN} \quad (8)$$

where the subscript $i$ represents the $i$-th row of a matrix, $W^{att} \in \mathbb{R}^{d \times d}$, $b^{att} \in \mathbb{R}^{1 \times d}$ are the learned parameters, A is the attention weights matrix and $\odot$ represents the element-wise multiplication. MDA computes a weighted score for each element, rather than row, in $V^{FFN}$. This way of weighted scores assignment enables MDA to encode more information than one single score used in traditional attention. We sum each row of $V^{FFN}$ with the attention weights to form the feature vector $s$. As described at the beginning of this section, two components share a same MDA, which means $V^{FFN}$ of each components is passed to a MDA with the same $W^{att}$ and $b^{att}$. We denote the resulting $s$ with respect two components as $s^{m_1}$ and $s^{m_2}$. Sharing the same MDA aims to select the information that connects the two entities, which is important to the PPI extraction task.

To further introduce the knowledge, we concatenated the relation embedding of the two protein entities and two feature vectors $s^{m_1}$ and $s^{m_2}$ to form the final feature representation $[s^{m_1}, s^{m_2}, e_r]$. Finally, $[s^{m_1}, s^{m_2}, e_r]$ is passed to a softmax layer to perform classification.

*3.2.3 Classification and Training*

We feed $[s^{m_1}, s^{m_2}, e_r]$ into a fully-connected layer followed by a softmax layer for relation classification.

$$p(y = j \mid I) = \text{softmax}(W_s[s^{m_1}, s^{m_2}, e_r] + b_s)$$

$$\hat{y} = \arg\max_{y \in [0,1]}(p(y = j \mid I)) \quad (9)$$

where $\hat{y}$ is our prediction, $W_s \in \mathbb{R}^{2 \times 3d}$ is a learned transformation matrix, $b_s$ is a learned bias vector, and $I$ is the training instances. The loss function is defined as follows:

$$J(\theta) = -\frac{1}{N}\sum_{i=1}^{N} \log p(y_i \mid I_i, \theta) \quad (10)$$

where $N$ is the number of labelled instances in the training set, $y_i$ is the golden label of the instance, $I_i$ is the $i$-th instance, and $\theta$ is the parameters of the entire model.

## 4 Experiments and Results

4.1 Dataset and Evaluation Metrics

**Dataset.** Experiments are conducted on the BioCreative VI Track 4 PPI extraction task corpus [25]. The organizers provide 597 training PubMed abstracts. The test set consists of 1,500 unannotated abstracts. Protein entities in the training and test sets are recognized by GNormPlus [38] toolkits[1], and normalized to Entrez Gene ID. We extract PPI relation triples from two knowledge bases, IntAct[2] and BioGrid[3], which have the same 45 kinds of relation types. Finally, 1,518,592 triples and 84,819 protein entities are obtained for knowledge representation training. All the protein entities in KBs are linked to the Entrez Gene IDs by using UniProt[4] [39] database.

---
[1] https://www.ncbi.nlm.nih.gov/CBBresearch/Lu/Demo/tmTools/download/GNormPlus/GNormPlusJava.zip
[2] https://www.ebi.ac.uk/intact/
[3] https://thebiogrid.org/

**Evaluation Metrics.** For PPI task, organizers employ two level evaluation:

Exact Match: All system predicted relations are checked against the manual annotated ones for correctness.

HomoloGene Match: All gene identifiers in the predicted relations and manually annotated data are mapped to common identifiers representing HomoloGene classes, then all predicted relations are checked for correctness.

We use Exact Match evaluation to measure the overall performance of our method. The evaluation of PPI extraction is reported by official evaluation toolkit[5], which adopts micro-averaged [40] Precision (P), Recall (R) and F1-score (F) based on Exact Match.

4.2 Experimental Setup

Word2Vec tool[6] [35] is used to pre-train word embeddings on the datasets (about 9,308MB, 27 million documents, 3.4 billion tokens and 4.2 million distinct words) downloaded from PubMed[7]. The dimensions of word, entity, and relation embeddings are all $d$=100. In DMHA we employ $h$=4 parallel attention layers, or heads. For each of these, the dimension of the projected $Q$, $K$ and $V$ is $d_{head}=d/h=25$. Thus, the dimensionality of input and output of FFN is 100 and we set the dimensionality of the inner-layer to 400. The model is trained by using Adadelta technique [41] with a learning rate 0.1 and a batch size 100. The whole framework is developed by PyTorch[8]. We release source code of our method on GitHub[9].

For knowledge representation learning, we initialize the entity embeddings with the averaged embeddings of words contained in entity mention and the relation embeddings with a normal distribution. The link of the TransE code is listed at footnote[10].

4.3 Effects of Architecture

In the experiment, the propose **KAN** is compared with the following baseline methods:

**CNN+MDA**: **KAN** adopts DMHA followed by FFN to capture long-range dependencies of the context. To compare with CNN, this variant of **KAN** replaces DMHA and FFN with a CNN. In the convolution layer, 100 feature maps with window size $k=\{3,4,5\}$ respectively are learned.

**LSTM+MDA**: To compare with LSTM, this variant of **KAN** replaces DMHA and FFN with a bidirectional LSTM, where the forward and backward LSTMs each have 100 hidden units.

The results of the two baselines are shown in Table 1. Unsurprisingly, **CNN+MDA** losses long-range dependency information and achieves a very low recall. **LSTM+MDA** has a higher recall than **CNN+MDA** due to the capacity of modeling variable-length sequences, but it has a relative low precision. Comparing with **CNN+MDA** and **LSTM+MDA**, **KAN** succeeds in capturing the long-rang knowledge-aware dependency information and provides the highest F1-score.

**Table 1.** Comparison with baselines.

| Architecture | Precision (%) | Recall (%) | F1-score (%) |
|---|---|---|---|
| **CNN+MDA** | 38.84 | 33.83 | 36.16 |
| **LSTM+MDA** | 36.57 | 35.93 | 36.24 |
| **KAN** | 38.07 | 37.28 | 37.67 |

To verify the effectiveness of each part of our **KAN**, we compare it with the three variants of **KAN**:

**KAN_SE**: **KAN_SE** only has one component of KAN, where the two protein embeddings are concatenated to each position to form $Q$ $\left\{[x_1^{m_1},e_1,e_2],[x_2^{m_1},e_1,e_2],\ldots,[x_l^{m_1},e_1,e_2],\ldots,[x_L^{m_1},e_1,e_2]\right\}$.

**w/o MDA**: Before the softmax layer, **KAN** adopts MDA to form the feature representation $[s^{m_1},s^{m_2}]$. Instead of MDA, this variant of **KAN** adopts the traditional attention mechanism which assigns a weighted score to each row, rather than element, of $V^{FFN}$.

---

[4] https://www.uniprot.org/
[5] https://github.com/ncbi-nlp/BC6PM
[6] https://code.google.com/p/word2vec/
[7] http://www.ncbi.nlm.nih.gov/pubmed/
[8] http://pytorch.org/
[9] https://github.com/zhuango/KAN
[10] https://github.com/thunlp/Fast-TransX

**w/o Mask**: In **KAN**, DMHA adopts a diagonal-disabled mask matrix $M$ to disable the attention of each token to itself. For this variant of **KAN**, we just drop the mask matrix.

**KAN_CNN*k***: This variant replaces the FFN in **KAN** with the CNN. CNN can extract more accurate relevance information by taking the consecutive features into account, which are expected to further improve the performance of **KAN**. We explores three windows size $k=\{3,4,5\}$ of CNN forming **KAN_CNN3**, **KAN_CNN4** and **KAN_CNN5**, respectively.

The results in Table 2 show that: (1) **KAN_SE** achieves a comparable results to **KAN**. The disadvantage of **KAN_SE** is the lower precision. We hypothesize that only one component of **KAN** would lose some specific clues about the PPI relation between two entities. After all, the parameters of **KAN_SE** is less than **KAN**, which reduces its learning ability. (2) From results of **KAN** and **w/o MDA**, we find that MDA decreases the precision, in return, it increases the recall and finally improves the F1-score. MDA selects more useful information than traditional attention, which makes it benefit from the trading off precision and recall. (3) The diagonal-disabled mask matrix in DMHA does slightly improve the F1-score. Disabling the attention of each token to itself is reasonable. (4) Replacing FFN with the CNN could improve the precision, leading to a higher F1-score than **KAN**. However, the computational complexity of convolutional operations affects the efficiency of **KAN_CNN*k***.

**Table 2.** Effects of components.

| Architecture | Precision (%) | Recall (%) | F1-score (%) |
| --- | --- | --- | --- |
| **KAN_SE** | 37.69 | 36.82 | 37.25 |
| **w/o MDA** | 38.43 | 36.71 | 37.55 |
| **w/o Mask** | 36.34 | 38.89 | 37.58 |
| **KAN_CNN3** | 42.05 | 34.98 | 38.19 |
| **KAN_CNN4** | 41.62 | 35.44 | 38.28 |
| **KAN_CNN5** | 38.88 | 37.63 | 38.25 |
| **KAN** | 38.07 | 37.28 | 37.67 |

To explore the effects of sharing parameters, **KAN** is compared with the following variants:

**KAN_SC**: In **KAN**, there are two components with respect to the two protein entities. Each component has its own set of parameters of DMHA and FFN. This variant totally shares the same parameters between two components (SC for short). That is, for two protein entities, **KAN_SC** adopts the same set of DMHA and FFN.

**KAN_DT**: **KAN** repeats DMHA and FFN multiple times to improve the performance, which means that they use the same set of parameters at each time. In contrast, this variant uses different DMHA and FFN at each time (DT for short), which means that they have different parameters at each time.

**KAN_DA**: **KAN** employs a shared MDA to filter information obtained from two components with respect to the two protein entities. In this variant, the two components do not share the MDA. That is to say, we use two different MDA to select information (DA for short).

**Table 3.** Effects of sharing parameters.

| Parameters Sharing | Precision (%) | Recall (%) | F1-score (%) |
| --- | --- | --- | --- |
| **KAN_SC** | 40.83 | 34.06 | 37.14 |
| **KAN_DT** | 34.65 | 40.39 | 37.30 |
| **KAN_DA** | 41.76 | 33.83 | 37.38 |
| **KAN** | 38.07 | 37.28 | 37.67 |

We report the results of these three variants in Table 3. **KAN_SC** would ignore some more specific information related to each entity. This may be the reason why **KAN_SC** performs worse, compared with **KAN**. **KAN_DT** achieves a higher recall. But it suffers with the lower precision and higher time complexity. **KAN_DA** neglects the communication between two entities and deems them as totally irrelevant things. The lower recall of **KAN_DA** illustrates that it loses the connection between two entities.

### 4.4 Effects of Prior Knowledge

This section explores the effects of prior knowledge in this section. Based on **KAN**, we remove some prior knowledge and get four variants of **KAN**.

**w/o Entity Embedding**: This variant uses no entity embedding at all, which means $Q$ is the same as $K$ and $V$ for DMHA.

**with Average Embedding**: In **KAN**, DMHA has three inputs denoted by $Q$, $K$, and $V$. $K$ and $V$ are from the same place, and the corresponding entity embedding is concat-enated to each row of *K* or *V* to form *Q*. The entity embeddings of **KAN** are learned from KBs through TransE model. This variant uses an average of constituting word embeddings of the entity mention to replace the TransE-based entity embedding.

**w/o Relation Embedding**: Before the softmax layer, **KAN** concatenates a relation embedding $e_r$ to the feature representation $[s^{m_1}, s^{m_2}]$ extracted by MDA. This variant of **KAN** discards the relation embedding and directly passes $[s^{m_1}, s^{m_2}]$ to the softmax layer.

**w/o KB**: Like two variants mentioned above, this variant of **KAN** not only uses an average of constituting word embeddings of the entity mention to replace the TransE-based entity embedding, but also discards the relation embedding.

**Table 4.** Effects of prior knowledge.

| Knowledge | Precision (%) | Recall (%) | F1-score (%) |
|---|---|---|---|
| w/o KB | 33.07 | 33.83 | 33.45 |
| w/o Relation Embedding | 33.84 | 35.79 | 34.79 |
| with Average Embedding | 35.36 | 37.51 | 36.40 |
| w/o Entity Embedding | 39.50 | 34.41 | 36.78 |
| KAN | 38.07 | 37.28 | 37.67 |

The results of the three variants of **KAN** are reported in Table 4. From the table, we could find that knowledge could dramatically improve the performance of PPI extraction.

(1) On the one hand, leveraging TransE-based entity embeddings brings 2.71% improvement in precision and further improves the F1-score by more than 1.27%. This indicates that TransE-based entity embeddings learned from the structural KBs are more effective than implicit word embeddings.

(2) On the other hand, although we employ relation embeddings in a very simple way, then increase both the precision and recall significantly. We believe that relation embeddings provide direct guidance for classification. After all, each kind of relation embeddings represents a specific relation contained in KBs.

(3) We also find that although using an average of constituting word embeddings of the entity mention is less effective than using the entity embedding learned through TransE, it does help **with Average Embedding** obtain more evidence about the PPI relation than **w/o Entity Embedding**, resulting in a higher recall than **w/o Entity Embedding**.

4.5 Performance relative to sequence length

To verify the effectiveness on extracting the long-range dependency information of our model, we explores how the recall and precision vary with sequence length. The candidate instances of test set are divided into five groups based on the sequence length range (<10, 10~15, 15~20, 20~30, >=30). We evaluate the recall and the precision of three models (**KAN_CNN4**, **KAN**, **MNM** (Zhou et al. [18])) for each group individually, as seen in **Fig. 2** and **Fig. 3**. From **Fig. 2**, we can conclude that **KAN** and **KAN_CNN4** outperform **MNM** in recall. And as the sequence length increased to 30 tokens, the difference of recall between our models and **MNM** gets bigger, which illustrates that our models can extract more PPIs on longer sequences than **MNM**. From **Fig. 3**, actually, the precision of **MNM** beats **KAN** in all five groups. After improving **KAN**, **KAN_CNN4** gets higher precision in all five groups than **KAN**, resulting in a higher precision than **MNM** on the whole test dataset.

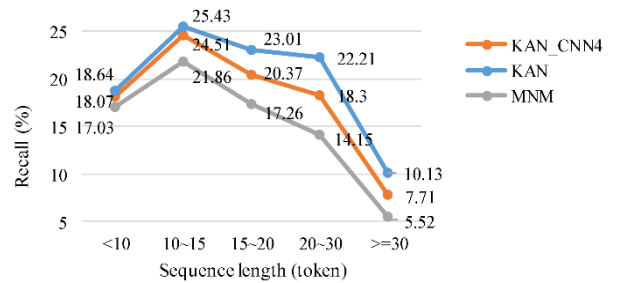

Fig.2. Recall relative to sequence length.

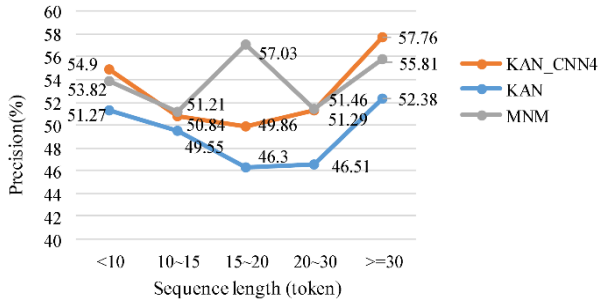

Fig.3. Precision relative to sequence length.

### 4.6 Distribution of entity relation triples

We split golden PPIs in test dataset into two groups in which triples are present and absent in KB and Training Dataset (KBTD), respectively. And the number of true positive PPIs predicted by **KAN** are shown in Table 5. With the help of prior knowledge in KB, 52% of in-KBTD PPIs in test set are extracted. Though we cannot obtain TransE-based relation embeddings of the not-in-KBTD PPIs, our model still extracts 18% of not-in-KBTD PPIs. Three examples are listed as follows:

(1) "*Immunoprecipitations were performed in myocytes expressing **PKCzeta** using PKC phospho-motif antibodies to determine the phosphorylation of **cTnI**, **cTnT**, tropomyosin, myosin-binding protein C, and desmin. (PMID: 17724026; Entity pairs: 5590, 7137 and 5590, 7139)*"

(2) "*The **LIMP-2** segment 145-288, comprising the nonsense mutations, contains a highly conserved coiled-coil domain, which we suggest determines **beta-GC** binding. (PMID: 19933215; Entity pair: 950, 2629)*"

(3) "*A **maspin** variant that has a point mutation of Arg(340) to Ala (Mas(R340A)) showed a significantly decreased affinity for **GST**. (PMID: 16049007; Entity pair: 5268, 373156)*"

All entity pairs in the above examples are not present in KBTD but recognized by **KAN**. All candidate instances from these examples are predicted as negative PPIs by **MNM** (Zhou et al. [18]) due to the complicated semantic and long range dependency between entity pair, while **KAN** correctly predicted them all. This illustrates that **KAN** is better than **MNM** on long sequences

**Table 5**. The distribution of entity relation triples. # means "the number of". Golden PPIs means PPIs actually exist in test dataset. Prediction means the true positive PPIs predicted by **KAN**. Percentage means that the percentage of Golden PPIs predicted by **KAN**. KBTD means knowledge base and training dataset.

|  | In KBTD | Not in KBTD |
| --- | --- | --- |
| #Golden PPIs | 496 | 373 |
| #Prediction | 258 | 66 |
| Percentage | 52% | 18% |

### 4.7 Computational complexity

For comparison of the computational complexity, we show the average time that models spend on one training epoch in Table 6. Not surprisingly, our models take more time than **MNM** because of the larger number of parameters. Nevertheless, our model allows for more parallelization, which could narrow the gap between **KAN** and **MNM**. After replacing the FFN with CNN, **KAN_CNN**$k$ ($k$ means the windows size) takes much more time than **KAN**. In general, **KAN** can give consideration to both computational complexity and performance.

**Table 6.** Time of one epoch (Sec).

| Model | Time(Sec) |
| --- | --- |
| **MNM** [18] | 21 |
| **LSTM+MDA** | 162 |
| **KAN_CNN3** | 142 |
| **KAN_CNN4** | 151 |
| **KAN_CNN5** | 163 |
| **KAN** | 78 |

## 5 Discussion

### 5.1 Comparison with Related Work

We compare our work with related work using both Exact Match evaluation measures in Table 7 and HomoloGene evaluation measures in Table 8. In order to make a fair comparison with every system and eliminate the influence of the accumulated errors introduced by different named entity recognition tools, all the systems are reported on the test dataset with the entity annotations recognized by GNormPlus [38] toolkits. We only compare Machine Learning-based (ML) methods without the post-processing rules, and divide these relevant systems into two groups: Machine Learning-based methods with or without additional resources, namely ML with KB and ML without KB.

Table 7. Comparison with related work (Exact Match evaluation)

| Methods | Related work | Precision (%) | Recall (%) | F1-score (%) |
|---|---|---|---|---|
| ML without KB | Chen et al.[3] | 34.49 | 32.87 | 33.66 |
| | Tran and Kavuluru [10] | 37.07 | 35.64 | 36.33 |
| | Wang et al.[15] | 9.81 | 50.59 | 16.43 |
| | Rios et al. [42] | 43.98 | 31.59 | 36.77 |
| ML with KB | Zhou et al. [18] | 40.32 | 32.37 | 35.91 |
| | **KAN_CNN4** | 41.62 | 35.44 | 38.28 |
| | **KAN_CNN4+Rule** | 35.19 | 40.74 | 37.76 |
| | **KAN** | 38.07 | 37.28 | 37.67 |
| | **KAN+Rule** | 33.24 | 41.54 | 36.93 |

Table 8. Comparison with related work (HomoloGene evaluation)

| Methods | Related work | Precision (%) | Recall (%) | F1-score (%) |
|---|---|---|---|---|
| ML without KB | Chen et al.[3] | 37.61 | 35.27 | 36.40 |
| | Tran and Kavuluru [10] | 40.07 | 38.63 | 39.33 |
| | Wang et al.[15] | 11.35 | 53.87 | 18.75 |
| ML with KB | Zhou et al. [18] | 42.47 | 34.22 | 37.90 |
| | **KAN_CNN4** | 43.78 | 37.46 | 40.37 |
| | **KAN_CNN4+Rule** | 37.57 | 43.70 | 40.41 |
| | **KAN** | 40.07 | 39.42 | 39.74 |
| | **KAN+Rule** | 35.45 | 44.51 | 39.47 |

In Table 7, among ML without KB methods, Rios et al. [42] achieve the best F1-score 36.77% in Exact Match evaluation measures. Rios et al. [42] take advantage of unlabeled data with neural adversarial learning, which gets 4% improvement.

In Table 8, among ML without KB methods, Tran and Kavuluru [10] obtain the best performance in Homolo-Gene evaluation measures. They adopt a CNN-based deep neural network for PPI extraction. Their system is simple but effective.

Chen et al. [3] use support vector machine with the graph kernel to extract PPI. Wang et al. [15] employ recurrent neural network to learn document representations for PPI extraction.

Among ML with KB methods, Zhou et al. [18] leverages prior knowledge about protein-protein pairs with memory networks. They achieve a poor recall because of losing long-rang dependency information. Our model explores the long-range dependency between words in a sequence to make more PPIs be recognized automatically. And the experimental results show that our model does effectively improve the recall without the post-processing rule that Zhou et al. [18] used. Therefore, **KAN** keeps the balance between the precision and the recall and achieves a state-of-the-art F1-score. After replacing the FFN in **KAN** with CNN, **KAN_CNN4** outperforms Zhou et al. [18] in precision. We also add the same post-processing rule that Zhou et al. [18] applied but the performance is decreased in general. Because our model has already recognized more PPIs, applying the post-processing would introduce more false positives, making a worse performance.

### 5.2 Attention visualization

To illustrate the effectiveness of attention, attention weights of DMHA in **KAN** and **KAN w/o KB** are visualized in the form of heat maps in Fig. 4 and Fig. 5, respectively. The example sentence is "*Our results suggest **MEK1** activated in differentiating myoblasts stimulates muscle differentiation by phosphorylating gene0 y156 which results in **MyoD** stabilization. (PMID: 21454680)*" where two protein entities are shown in bold type. As mentioned in methods section, the DMHA layer and FFN layer in **KAN** are repeated twice. Thus, we visualize the attention weights of the last or the second DMHA layer. For both Fig. 4 and Fig. 5, there are two attention weights matrixes. The left one corresponds to entity '***MEK1***' and the right one corresponds to entity '***MyoD***'. Note that DMHA performs four-head self-attention and the heat maps are generated according to the averaged attention weights of four self-attentions. We find that **KAN** learns to pay more attention to the words '*stabilization*' and '*stimulates*'. In the example sentence, we observe that '*stabilization*' actually describes the interaction between two entities. While in Fig. 5, the words that have highest weights are '*in*' and '*stimulates*'. **KAN w/o KB** fails to capture the key information from the word '*stabilization*'.

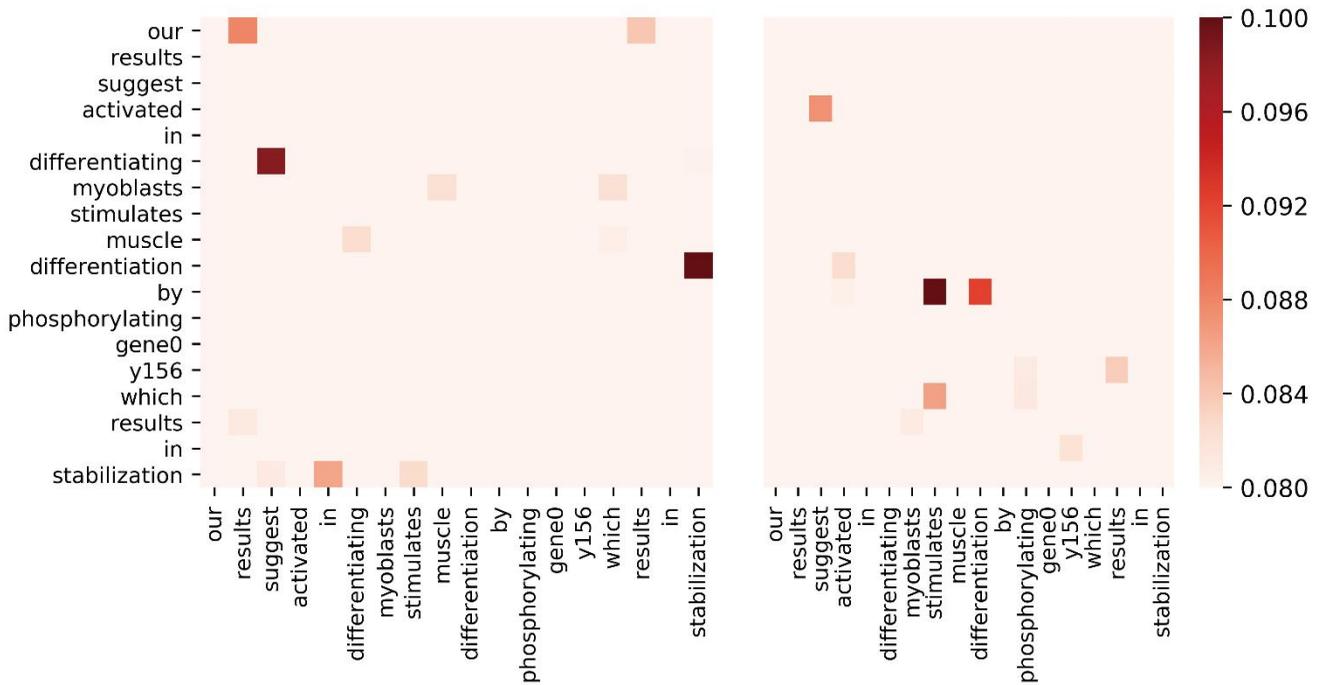

Fig.4. Visualization of attention weights by a heat map of **KAN**. The left panel corresponds to entity '***MEK1***' and the right panel corresponds to entity '***MyoD***'. Deeper color means higher weight.

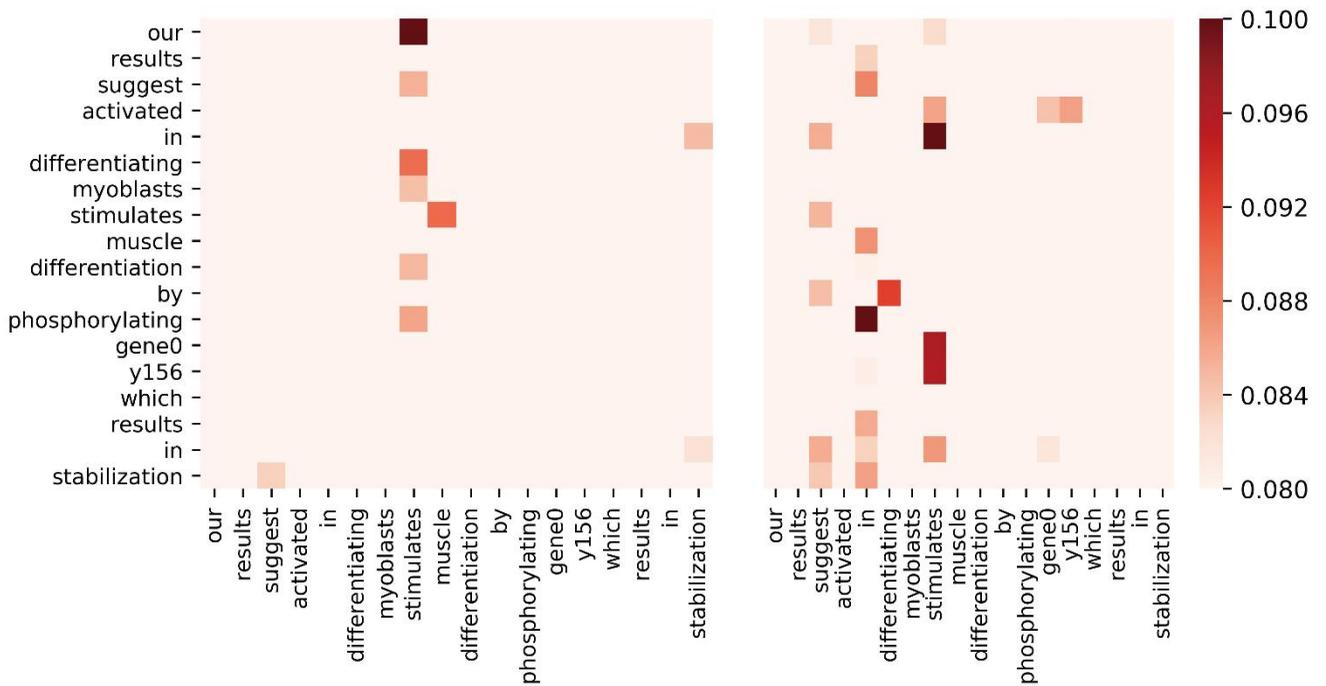

Fig.5. Visualization of attention weights by a heat map of **KAN w/o KB**. The left panel corresponds to entity '***MEK1***' and the right panel corresponds to entity '***MyoD***'. Deeper color means higher weight.

### 5.3 Error analysis

We perform an error analysis of the results of **KAN** to detect the origins of false positives (FPs) and false negatives (FNs) errors. All the FPs are from incorrect classification, for example, "*Mutation of Ser 193 to Ala also abolishes the ability of **C/EBPalpha** to cause growth arrest because of a lack of interactions with **cdk2** and E2F-Rb complexes. (PMID: 15107404)*", where two protein entities are shown in bold type. We can see that although there is a key word "*interactions*" in the sentence, two entities do not participate in a PPI relation. **KAN** fails to identify the negation keywords "abolish" or "a lack of".

FNs come from three factors, false negative entity caused by the GNormPlus toolkits, pre-processing rules and incorrect classification. GNormPlus fails to recognize part of entities in the test dataset, which directly leads to 386 FNs with a proportion of 70.44%. We filter out protein pairs distributed across more than two sentences by pre-processing rules in our system, resulting in 68 FNs with a proportion of 12.41%. Not surprisingly, **KAN** incorrectly classifies 91 interacting protein pairs as negative, for example, "*Here, we identified **S100A10** as the first auxiliary protein of these epithelial Ca(2+) channels using yeast two-hybrid and GST pull-down assays. This S100 protein forms a heterotetrameric complex with annexin 2 and associates specifically with the conserved sequence VATTV located in the C-terminal tail of **TRPV5** and TRPV6. (PMID: 12660155)*", where the two protein entities are shown in bold type. These sentences describe the relation of two entities in a very complicated semantic environment. The interaction between "***S100A10***" and "***TRPV5***" is mediated by the coreferential relationship. "***S100***" in the second sentence refers to the "***S100A10***" in the first sentence, and the second sentence conveys the interaction relationship. Detecting PPI relation in such sentences is beyond the capacity of **KAN**.

## 6 Conclusion

This paper develops a knowledge-aware attention network for PPI extraction task. The two level DMHA mechanism and MDA mechanism are adopted for learning long-range dependencies to make full use of the word context information. Further, prior knowledge are encoded into knowledge representations through TransE model, and incorporated into KAN. Experimental results show that both context representations and knowledge representations are effective in improving PPI extraction performance. Our KAN obtains the state-of-the-art results on BioCreative VI PPI dataset.

## References

[1] H. Hermjakob, L. Montecchi-Palazzi, C. Lewington, et al., "IntAct: an opensource molecular interaction database." *Nucleic Acids Res.*, 32, 452D–4455, 2004.

[2] C. Stark, B. Breitkreutz, T. Reguly et al., "BioGRID: a general repository for interaction datasets." *Nucleic Acids Res.*, 34, D535–D539, 2006.

[3] Q. Chen, N. C. Panyam, A. Elangovan et al., "Document triage and relation extraction for protein-protein interactions affected by mutations." In: *Proceedings of the 2017 Workshop on BioCreatice VI, Washington, DC*. pp. 103–106, 2017.

[4] I. Segurabedmar, P. Martinez, C. D. Pablesanchez., "Using a shallow linguistic kernel for drug-drug interaction extraction." *Journal of Biomedical informatics*, Volume 44 Issue 5, pp. 789-804, 2011.

[5] S. Kim, H. Liu, L. Yeganova. W. J. Wilbur, "Extracting drug-drug interactions from literature using a rich feature-based linear kernel approach." *Journal of Biomedical informatics*, Volume 55, pp. 23-30, 2015.

[6] L. Qian and G. Zhou, "Tree kernel-based protein-protein interaction extraction from biomedical literature." *J. Biomed. Inform.*, 45, 535–543, 2012.

[7] T. T. Phan and T. Ohkawa, "Protein-protein interaction extraction with feature selection by evaluating contribution levels of groups consisting of related features." *BMC Bioinformatics*, 17, 518–246543, 2016.

[8] D. Zeng, K. Liu, S. Lai et al., "Relation classification via convolutional deep neural network." In: *Proceedings of COLING 2014, the 25th International Conference on Computational Linguistics, Dublin*. pp. 2335–2344, 2014.

[9] Y. Lecun, L. Bottou, Y. Bengio et al., "Gradient-based learning applied to document recognition." In: *Proceedings of IEEE*, 86, 2278–2324, 1998.

[10] T. Tran and R. Kavuluru, "An end-to-end deep learning architecture for extracting protein-protein interactions affected by genetic mutations" *Database*, bay092, 2018.

[11] J. Schmidhuber, "Learning complex, extended sequences using the principle of history compression." *Neural Computation*, 4, 234–242, 1992.

[12] S. Hochreiter and J. Schmidhuber, "Long short-term memory." *Neural Computation*, 9, 1735–1780, 1997.

[13] J. Weston, S. Chopra, and A. Bordes, "Memory networks." *arXiv Preprint* arXiv: 1410.3916, 2014.

[14] S. Sukhbaatar, A. Szlam, J. Weston et al., "End-to-end memory networks." In: *Proceedings of the Twenty-Ninth Annual Conference on Neural Information Processing Systems, Montreal*. pp. 2440–2448, 2015.

[15] Y. Wang, F. Shen, R. K. Elayavilli et al., "MayoNLP at the BioCreative VI PM Track: Entity enhanced Hierarchical Attention Neural Networks for Mining Protein Interactions from Biomedical Text." In: *Proceedings of the 2017 Workshop on BioCreatice VI, Washington, DC*. pp. 127–130, 2017.

[16] H. Zhou, H. Deng, L. Chen et al., "Exploiting syntactic and semantics information for chemical–disease relation extraction." *Database*, baw048, 2016.

[17] S. K. Sahu and A. Anand, "Drug-drug interaction extraction from biomedical texts using long short-term memory network" *Journal of Biomedical informatics*, Volume 86, pp. 15-24, 2018.

[18] H. Zhou, Z. Liu, S. Ning and Y. Yang, "Leveraging Prior Knowledge for Protein-Protein Interaction Extraction with Memory Network." *Database*, bay071, 2018.

[19] Y. Shen, and X. Huang, "Attention-based convolutional neural network for semantic relation extraction." In: *Proceedings of COLING 2016, the 26th International Conference on Computational Linguistics, Osaka*. pp. 2526–2536, 2016.

[20] A. Vaswani, N. Shazeer, N. Parmar et al., "Attention is All you Need." In *Proceedings of the Thirtieth Annual Conference on Neural Information Processing Systems, Long Beach Convention Center, Long Beach*, pp. 6000–6010, 2017.

[21] T. Shen, T. Zhou, G. Long et al., "DiSAN: Directional Self-Attention Network for RNN/CNN-free Language Understanding." In: *Proceedings of the Thirty-Second AAAI Conference on Artificial Intelligence, New Orleans, Louisiana*, 2018.

[22] F. Alam, A. Corazza, A. Lavelli et al., "A knowledge-poor approach to chemical-disease relation extraction." Database, 2016, baw071, 2016.

[23] E. Pons, F. B. Becker, A. S. Akhondi et al., "Extraction of chemical-induced diseases using prior knowledge and textual information." *Database*, baw046, 2016.

[24] A. Bordes, N. Usunier, A. Garciaduran et al., "Translating embeddings for modeling multi-relational data." In: *Proceedings of the Twenty-Seventh Annual Conference on Neural Information Processing Systems, Lake Tahoe*. pp. 2787–2795, 2013.

[25] I. R. Doğan, A. Chatr-Aryamontri, H. C. Wei et al., "The BioCreative VI precision medicine track corpus." In: *Proceedings of the 2017 Workshop on BioCreatice VI, Washington*. pp. 88–93, 2017a.

[26] Z. Wang, J. Zhang, J. Feng et al., "Knowledge graph embedding by translating on hyperplanes." In: *Proceedings of the Twenty-Eighth AAAI Conference on Artificial Intelligence, Quebec*. pp. 1112–1119, 2014.

[27] Y. Lin, Z. Liu, M. Sun et al., "Learning entity and relation embeddings for knowledge graph completion." In: *Proceedings of the Twenty-Ninth AAAI Conference on Artificial Intelligence, Texas*. pp. 2181–2187, 2015.

[28] G. Ji, S. He, L. Xu et al., "Knowledge Graph Embedding via Dynamic Mapping Matrix." In: *Proceedings of the 53rd Annual Meeting of the Association for Computational Linguistics and the 7th International Joint Conference on Natural Language Processing, Beijing, China's*, pp. 687–696, 2015.

[29] G. Ji, K. Liu, S. He et al., "Knowledge graph completion with adaptive sparse transfer matrix." In: *Proceedings of the Thirty-Second AAAI Conference on Artificial Intelligence, New Orleans, Louisiana*. pp. 985-991, 2016.

[30] J. Cheng, L. Dong, and M. Lapata, "Long short-term memory-networks for machine reading." *arXiv preprint arXiv: 1601.06733*, 2016.

[31] A. Parikh, O. Täckström, D. Das, and J. Uszkoreit, "A decomposable attention model." in *Proceedings of the 2016 conference on empirical methods in natural language processing (EMNLP)*, pp. 2249-2255, 2016.

[32] R. Paulus, C. Xiong, and R. Socher. "A deep reinforced model for abstractive summarization." *arXiv preprint arXiv: 1705.04304*, 2017.

[33] Z. Lin, M. Feng, C. N. Santos, M. Yu, B. Xiang, B. Zhou, and Y. Bengio, "A structured self-attentive sentence embedding." *arXiv preprint arXiv:1703.03130*, 2017.

[34] I. R. Doğan, S. Kim, A. Chatr-Aryamontri et al., "Overview of the BioCreative VI precision medicine track." In: *Proceedings of the 2017 Workshop on BioCreatice VI, Washington, DC*. pp. 83–87, 2017b.

[35] T. Mikolov, I. Sutskever, K. Chen et al., "Distributed representations of words and phrases and their compositionality." In: *Proceedings of the Twenty-Seventh Annual Conference on 40 Neural Information Processing Systems, Lake Tahoe*. pp. 3111–3119, 2013.

[36] K. He, X. Zhang, S. Ren, and J. Sun, "Deep residual learning for image recognition." In: *Proceedings of the IEEE Conference on Computer Vision and Pattern Recognition*, pp. 770–778, 2016.

[37] J. L. Ba, J. R. Kiros, and G. E. Hinton, "Layer normalization." *arXiv preprint arXiv:1607.06450*, 2016.

[38] C. Wei, H. Kao, and Z. Lu, "GNormPlus: an integrative approach for tagging genes, gene families, and protein domains." *BioMed Res. Int.*, 1–7, 2015.

[39] UniProt Consortium, "UniProt: a hub for protein information." *Nucleic Acids Res.*, 43, D204–D212, 2015.

[40] M. Krallinger, A. Morgan, L. Smith et al., "Overview of the protein-protein interaction annotation extraction task of BioCreative II." *Genome Biol.*, 9, S1–S19, 2008.

[41] D. M. Zeiler, "ADADELTA: an adaptive learning rate method." *CoRR* abs/1212.5701, 2012.

[42] A. Rios, R. Kavuluru and Z. Lu, "Generalizing biomedical relation classification with neural adversarial domain adaptation." *Bioinformatics*, bty190, 2018.